%% file: main.tex
\documentclass[twoside]{article}
\usepackage[accepted]{aistats2018}

\usepackage[utf8]{inputenc} 
\usepackage[T1]{fontenc}    
\usepackage{hyperref}       
\usepackage{url}            
\usepackage{booktabs}       
\usepackage{amsfonts}       
\usepackage{nicefrac}       
\usepackage{microtype}      

\usepackage{times}
\usepackage{graphicx} 
\usepackage{natbib}
\usepackage{algorithm}
\usepackage{algorithmic}
\usepackage{amsmath,amssymb,dsfont}

\usepackage{amssymb,dsfont}
\usepackage{caption}

\usepackage{tikz}
\usetikzlibrary{bayesnet}

\newcommand{\distthetaalpha}{0.5}
\newcommand{\distztheta}{0.5}
\newcommand{\distwz}{0.5}
\newcommand{\distbetatau}{0.8}
\newcommand{\distbetaw}{1.0}
\newcommand{\distww}{0.5}
\newcommand{\tauleftshift}{7}

\newcommand{\dd}{\mathrm{d}}

\newcommand{\EE}{\mathbb{E}}

\newcommand{\uptoconst}{\mathrel{\overset{\makebox[0pt]{\mbox{\normalfont\small\sffamily c}}}{=}}}

\DeclareMathOperator{\tr}{tr}

\begin{document}

\twocolumn[

\aistatstitle{Scalable Generalized Dynamic Topic Models}

\aistatsauthor{Patrick J\"ahnichen$^{\;*\;1}$ \And Florian Wenzel$^{\;*\;1\;2}$ \And Marius Kloft$^{\;1\;2}$ \And Stephan Mandt$^{\;3}$}
\aistatsaddress{$^1$Humboldt-Universit\"at zu Berlin, Germany \And $^2$TU Kaiserslautern, Germany \And $^{3}$Disney Research, Los Angeles, USA} ]
\runningauthor{Patrick J\"ahnichen, Florian Wenzel, Marius Kloft, Stephan Mandt}

\begin{abstract} 
Dynamic topic models (DTMs) model the evolution of prevalent themes in literature, online media, and other forms of text over time. DTMs assume that word co-occurrence statistics change continuously and therefore impose continuous stochastic process priors on their model parameters. These dynamical priors make inference much harder than in regular topic models, and also limit scalability. In this paper, we present several new results around DTMs. First, we extend the class of tractable priors from Wiener processes to the generic class of Gaussian processes (GPs). This allows us to explore topics that develop smoothly over time, that have a long-term memory or are temporally concentrated (for event detection). Second, we show how to perform scalable approximate inference in these models based on ideas around stochastic variational inference and sparse Gaussian processes. This way we can train a rich family of DTMs to massive data. Our experiments on several large-scale datasets show that our generalized model allows us to find interesting patterns that were not accessible by previous approaches.
\end{abstract}

\input{intro}

\input{related-work}
\input{model}

\input{inference}

\input{experiments}

\input{conclusion}

\section*{Acknowledgements} 
This work was partly funded by the German
Research Foundation (DFG) award KL 2698/2-1 and
the Federal Ministry of Science and Education (BMBF) awards 031L0023A and 031B0187B.

\bibliographystyle{apa}
\bibliography{bib}

\begin{appendix}
	\include{appendix}
\end{appendix}

\end{document}

%% file: intro.tex
\section{Introduction}
\label{intro}

Probabilistic topic models help us to organize and browse large collections of documents \citep{blei2012probabilistic}. 
Topic models have been successfully applied in information retrieval \citep{McCallum:2004p18,Wang:2007p51,charlin2013toronto}, computational biology~\citep{pritchard2000inference,gopalan2016scaling}, recommendation systems~\citep{wang2011collaborative}, and computer vision~\citep{fei2005bayesian,chong2009simultaneous}. Topic models assume that all words in a document were independently drawn from a finite set of probability distributions over words, termed the 'topics'. This way, every document is a mixture of topics. The limitation is that this approach assumes that topics are static.

Topics change over time. To provide some intuition, consider the example of the topic \emph{technology} when training topic models on historical articles~\footnote{
Example from David Blei's tutorial slides on topic modeling, \url{http://www.cs.columbia.edu/~blei/talks/Blei_ICML_2012.pdf}}. Restricting the corpus to articles around 1900, we find words such as \emph{engine}, \emph{electricity}, and \emph{wire} to be mainly associated with this topic. For modern articles, we may find \emph{devices}, \emph{gates}, and \emph{silicon} among the top words. In applications as this, we want to be able to associate documents with similar topic proportions with each other over large time spans. But at the same time, we want to allow topics to  'modernize', meaning to dynamically adjust their vocabulary. This is achieved in dynamic topic models (DTMs)~\citep{Blei:2006wj,Wang:2006p724,Wang2008continuous}. DTMs model the evolution of topics as a continuous Wiener process. This dynamic prior determines how strongly topics may change their vocabulary. This way, DTMs share statistical strengths over all times, while giving the topics enough flexibility to change.

Current formulations of dynamic topic models are subject to the major limitation that they are restricted to a particular type of stochastic process for the latent topical dynamics, namely Wiener processes. This formulation does not allow us to analyze long-term effects, events, or other more complicated temporal dependencies. 
Second, relying on the forward-backward algorithm, they lack scalability. If the data are distributed across many different time-stamps, they require a full pass through the data in every iteration. This lack of scalability may be the reason why  DTMs have been much less used in large-scale scientific or industrial applications than their static counterparts.
In this paper, we generalize dynamic topic models in two ways: first we extend the class of tractable priors from Wiener processes to the more general class of Gaussian processes. Second, we derive a scalable approximate Bayesian inference algorithm based on inducing points. This allows us to apply our model to contemporary large text collections. In more detail, our main contributions are as follows:  

\begin{itemize}
\item We formulate DTMs in terms of latent Gaussian process priors on topic evolution. 
This opens a wealth of possibilities for new models in which the topics display different types of temporal (or even spatial) correlations. 
Going beyond the typical Wiener processes, we analyze 
Ornstein-Uhlenbeck processes for event detection, 
Gaussian processes with Cauchy kernels (for long-term memory effects) 
and squared exponential kernels (for rather short-term memory effects). 
\item We derive a scalable variational inference algorithm for this new model class. 
Our approach relies on inducing points for Gaussian process inference~\citep{Snelson:2006vi,Titsias:2009vf,Hensman:2013tn}. 
All natural gradients are given in closed-form and do not rely on numerical optimization or sampling approaches. 
Natural gradients have the advantage that they are invariant to reparameterization of the variational family \citep{infogeom,naturalgrad} and provide effective second-order optimization updates~\citep{JMLR:v14:hoffman13a, wenzel2018GP}.
While a naive implementation would scale cubically in the number of time stamps, our approach scales cubically in the number of inducing points, which is typically much smaller. 
\item In our experiments, we investigate dynamic topics using different kernels. These new priors allow us to find patterns which were not accessible before. For instance, we filter time-localized topics in a set of speeches on the State of the Union and in news articles as published in the New York Times.
\end{itemize}

This paper is organized as follows.
In section~\ref{sec:related_work} we discuss related work. We describe the novel generalized dynamic topic model in section~\ref{sub:topic_model} and present an efficient variational inference algorithm for our model in section~\ref{sec:inference}. Section~\ref{sec:experiments} concludes with experiments.
For implementation details, we refer readers to the website of the first author of this paper\footnote{\url{https://patrickjae.github.io}}.

%% file: related-work.tex
\section{Background and Related Work}
\label{sec:related_work}
We connect to dynamic and correlated topic models, sparse GPs and stochastic variational inference (SVI).

\paragraph{Dynamic Topic Models.} DTMs form the basis of our approach. While~\citet{Blei:2006wj} originally proposed a model with equidistant time slices,~\citet{Wang2008continuous} extended the approach to continuous time. Both rely on a latent Wiener process and use the forward-backward algorithm for learning, which requires full passes through the data in every iteration if the number of time stamps is comparable with the total number of documents. \citet{Wang:2006p724} proposed a different approach where time is an observed variable with some prior over a finite time interval. While in principle being scalable, the resulting topics are non-smooth. Finally,~\citet{bhadury2016scaling} proposed a new approach for learning in topic models based on stochastic gradient MCMC~\citep{Welling:2011ws,Mandt:2016vz}. Their approach similarly is restricted to latent Wiener processes.

\paragraph{Correlated and GP Topic Models.} This class of modified static topic models breaks the independence assumptions of the per-document topic proportions. Instead, the topic proportions are jointly drawn from some prior which induces correlations~\citep{Blei:2007vt}. If this prior is a Gaussian process, this leads to the kernel topic model~\citep{hennig2012kernel} or Gaussian process topic model~\citep{agovic2012gaussian}. Note that both approaches assume that the topics themselves are static and only the topic proportions change. In contrast, we treat the proportions as independent and identically distributed (iid) and impose dynamics on the topics themselves. None of these models have been formulated in a scalable manner.

\paragraph{Stochastic Variational Inference and sparse GPs.} Our algorithm builds on stochastic variational inference (SVI)~\citep{JMLR:v14:hoffman13a}, which combines variational inference with stochastic optimization.
SVI can normally only applied if the data are iid conditioned on a global set of paramaters, which is an assumption that is typically broken in Gaussian process modelling setups.
\citet{Hensman:2013tn,Hensman:2012tx} have shown that one can derive a tractable lower bound to the marginal likelihood of the data that allows for data subsampling. This so-called inducing point or sparse approach dates back to earlier work by~\citet{Titsias:2009vf, Snelson:2006vi} and~\citet{csato2002sparse} and has been successfully applied to a variaty of GP models~\citep[e.g.][]{Hensman:2015uj,b-svm}. 
None of this work has been applied in the context of topic models.

%% file: model.tex
\section{Generalized Dynamic Topic Models}
\label{sub:topic_model}
Dynamic topic models are mixed-membership bag-of-words models
which allow their mixture components---the topics---to drift over time. 
This allows to dynamically fade-in new words, and fade-out old words which loose their semantic significance in a topic.
In the classic DTM this continuity is achieved by imposing a Wiener process prior on the topic matrices~\citep{Blei:2006wj,Wang2008continuous} (see also~\citep{bamler2017dynamic} for a related approach for word embeddings). 

In this paper, we propose Gaussian processes as priors on the topic matrices. Since the Wiener process is a specific type of GP, our approach is a strict generalization of dynamic topic models but covers a much richer class of dynamics.
We introduce the generalized dynamic topic model in section~\ref{sec:DTM1} and present a scalable version in section~\ref{sec:DTM2}.

\subsection{Generalized DTMs}
\label{sec:DTM1}
For what follows, we borrow notation from the topic modeling literature~\citep{Blei:2003p5}. We assume that we observe a corpus of $D$ documents, each of which is associated with a time stamp $\tau_{t_d}$ with index $t_d \in \{1,\ldots,T\}$. 
For a simpler notation we denote the number of words in a document as $N$.
For a given document $d$ with time index $t$, let $w_{d1},\cdots,w_{dN}$ be the
words it contains,
$\theta_{d}$ be a $K$-vector of topic proportions and $z_{dn}$ 
the assignment of word $w_{dn}$ to a topic. The model consists of $K$ time dynamic topics whereby $\beta_{k\cdot t}$ denotes a topic's  $V$-dimensional distribution over the vocabulary at time $t$.

\begin{figure}
\centering
    \input{plate_figures/gdtm_plate_vertical}
        \caption{The generalized dynamic topic model.}
        \label{fig:gdtm}
\end{figure}
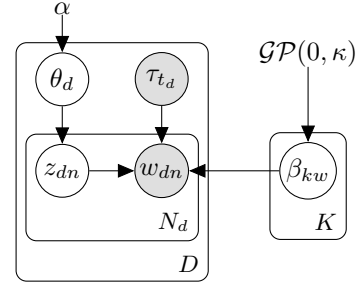

Our model exhibits the following joint distribution:
\begin{align}
    p(w,z,\theta,\beta) &= p(\beta)\prod_{t=1}^T \prod_{k=1}^K p(w_{t},z_{t},\theta | \pi(\beta_{k\cdot t})).
\end{align}
The function $\pi(\cdot)$ is the softmax function which normalizes the topic $\beta_{k\cdot t}$ over the vocabulary indices. The remaining likelihood, 
\begin{align*}
    p(w_t,&z_t,\theta | \pi(\beta_{\cdot \cdot t})) =\\ &\prod_{d:t_d=t}p(\theta_d) \prod_n p(w_{dn}|\pi(\beta_{z_{dn}\cdot t})) p(z_{dn}|\theta_d),
\end{align*}
is just a regular LDA model (at time $t$), where $p(w_{dn}|\pi(\beta_{z_{dn}\cdot t}))={\rm Mult}(\pi(\beta_{z_{dn}\cdot t}))$, $p(z_{dn}|\theta_d) = {\rm Mult}(\theta_d)$, $p(\theta_d) = {\rm Dir}(\alpha)$. The graphical model is shown in Figure~\ref{fig:gdtm}.

The distinctive feature of dynamic topic models is their dynamic prior $p(\beta)$. In our model each of the $V$ words out of $K$ topics is a latent function over time, drawn from a GP with kernel function $\kappa$. This GP is observed at times $\tau_1,\cdots,\tau_T$ and can thus be described as a $T$-dimensional multivariate normal distribution\footnote{We call attention to the slight overloading of notation: a plain $K$ always is the number of topics, using subscripts or a tilde it denotes a kernel/covariance matrix.}:
\begin{align}
\beta_{kw\cdot} &\sim {\rm GP}(0, \kappa) \Leftrightarrow \beta_{kw,1:T} \sim \mathcal N_T(0,K_{TT}),\label{eq:beta_prior}\\
K_{\tau, \tau^\prime} &= \kappa(\tau, \tau^\prime),\quad \tau, \tau^\prime \in \{\tau_1, \cdots, \tau_T\}.
\end{align}
Using a Wiener kernel function in our model results in the classic DTM of \citet{Wang2008continuous}.
However, due to the model's flexibility we can model \textit{any} stochastic process that falls into the class of GPs by simply altering the covariance function $\kappa$.
As an aside, this setup not only covers the dynamic setup, but also allows for incorporating other types of meta data as e.g. spatial modeling if the text documents are associated with location coordinates.

In this paper, we focus on the time-specific setup. In more detail, we consider several different kernels commonly used for time-series modeling \citep{Roberts:XHzSIn0E}.

\begin{itemize}
\item \textbf{Wiener kernels}, $\kappa_\text{Wie}(\tau,\tau') = \sigma^2 \min(\tau,\tau')$. Using a Wiener kernel (Brownian motion kernel) in our model recovers the typical DTM setup. This serves as our baseline. 

\item \textbf{Ornstein-Uhlenbeck kernels}, $\kappa_\text{OU}(\tau,\tau') = \sigma^2 \exp \left(-\frac{|\tau-\tau'|}{l} \right)$.
The Ornstein-Uhlenbeck (OU) process is essentially a Wiener process in the presence of a mean-reverting force which pulls the process state back to its mean and thus acts like a regularizer. An effect of this is that topics may die-off and other topics may dynamically emerge (using a zero-mean process). As we show in our experiments, this leads to temporally localized changes in topics.


\item \textbf{Squared exponential kernels}, $\kappa_\text{SE}(\tau,\tau') = \sigma^2 \exp \left(-\frac{(\tau-\tau')^2}{2l^2} \right)$.
Squared exponential (SE) kernels have the property that the resulting trajectories are smoother compared to Wiener kernels. The resulting prior functions are infinitely often differentiable. The exponential decay of the temporal correlations
leads to memory effects that can be parameterized by the kernel's length scale  $l$.
With a suitable chosen $l$ this allows for temporally localized topics.


\item \textbf{Cauchy Kernels}, $\kappa_\text{Cau}(\tau,\tau') = \sigma^2 \left( 1 + \frac{(\tau-\tau')^2}{l^2} \right)^{-1}$.
Cauchy kernels are constructed similarly as SE kernels, but instead of using the Gaussian density one uses a Cauchy density. This kernel has long-range memory, which means that temporal correlations decay not exponentially but polynomially, which in some cases is more realistic.
\end{itemize}

Note that \emph{any} additive or multiplicative combination of covariance functions again results in a valid covariance function again and so can similarly be used. This adds considerable to the flexibility of the proposed prior.

We again stress that all these kernels use the same inference algorithm. The problem is that a naive implementation would scale cubicly in the number of time stamps. We therefore propose a more efficient version based on the concept of sparse GPs.

\subsection{Sparse DTMs}
\label{sec:DTM2}

The bottleneck of inference in the model introduced in section~\ref{sec:DTM1} is the inversion of the $T\times T$ kernel matrix $K_{TT}$. One solution is to bin time stamps into groups, thus artificially reducing $T$. But this way we loose valuable information, especially when the number of distinct time stamps is comparable to the number of observed documents themselves, i.e. $T \approx D$. 

Instead, we present a scalable version of the generalized dynamic topic model based on inducing points~\citep{Hensman:2013tn}. This is a low-rank approximation to the $T$-dimensional GPs based on $\hat T$ artificial time stamps (inducing points) where $\hat T \ll T$.
The inversion necessary for inference is only based on the $\hat T\times\hat T$ covariance matrix of the approximating GP and can therefore be computed efficiently.

Following~\citet{Hensman:2013tn}, let $K_{TT}$ be the kernel evaluated at all training points (i.e. the full rank kernel as in \eqref{eq:beta_prior}), $K_{\hat{T}\hat{T}}$ the kernel evaluated at inducing points, and $K_{T\hat{T}}$ and $K_{\hat{T}T}$ be kernels evaluated in-between these sets of points. Furthermore, let $u$ be a $\hat{T}$-dimensional variable. We make use of the following Gaussian integral:
\begin{align}
    {\cal N}(0,K_{TT}) = \int {\cal N}(K_{T\hat T}K_{\hat T\hat T}^{-1}&u,\tilde K){\cal N}(u; 0,K_{\hat{T}\hat{T}}) \dd u,
\end{align}
where $\tilde K = K_{TT} - K_{T\hat T}K_{\hat T\hat T}^{-1}K_{\hat TT}$.
Thus, we introduce latent auxiliary variables $u_{kw}$ for every $\beta_{kw}$ such that the resulting marginal distribution of $\beta_{kw}$ does not change (when integrating over $u_{kw}$). Defining $p(u_{kw}) = \mathcal N(0,K_{\hat T \hat T})$, we obtain 
\begin{equation}
    p(\beta_{kw}|u_{kw}) = \mathcal N(K_{T\hat T}K_{\hat T\hat T}^{-1}u_{kw}, \tilde K),
    \label{eq:lowrank_approx}
\end{equation}
and perform approximate inference over $u$. Also note that conditioning of GPs involves inversion of the kernel matrix. In our approach, inverting a $T\times T$ matrix is now replaced by inverting one of size $\hat{T}\times \hat{T}$. 

The augmented joint distribution is 
\begin{align}
	p(\beta, w, z, \theta, u ) = p(w | \beta,z) p(z|\theta) p(\beta | u) p(u). \label{eq:augmented_joint}
\end{align}
This summarizes our model (the discussion of marginalizing over $\beta$ is deferred to the next section).
Next, we present details about the inference procedure. Readers primarily interested in experimental results may therefore skip section~\ref{sec:inference} and continue with section~\ref{sec:experiments}.

%% file: plate_figures/gdtm_plate_vertical.tex
\begin{tikzpicture}[x=1.2cm,y=1cm]

  \node[const]						(GP)      	{$\mathcal{GP}(0, \kappa)$}; %

  \node[latent, below=of GP] 		(beta)		{$\beta_{kw}$} ; %
  \node[obs, left=\distbetaw of beta]	 	(w)   		{$w_{dn}$} ; %
  \node[latent, left=\distwz of w]       (z)       {$z_{dn}$} ; %
  \node[latent, above=\distztheta of z]       (theta)   {$\theta_d$}; %
  \node[const, above=\distthetaalpha of theta]   (atheta)  {$\alpha$};


  \node[obs, above=\distztheta of w] 		(t)	   		{$\tau_{t_d}$} ; %

  \edge {atheta} {theta}
  \edge {theta} {z}
  \edge {z} {w}
  \edge {t} {w}
  \edge {beta} {w}
  \edge {GP} {beta}

  \plate {plate_doc} { 
    (z)(w) %
  } {$N_d$}; %
  \plate {plate_corpus} { %
    (plate_doc) (t) (theta)%
  } {$D$} ; %
  \plate {plate_topics} { %
    (beta)
  } {$K$}; %

\end{tikzpicture}

%% file: inference.tex
\section{Inference}
\label{sec:inference}
In Bayesian latent variable models such as DTMs, our goal is to compute the posterior distribution over the latent variables. 
This quantity is intractable and we have to resort to approximate methods. We use variational inference, which maps the inference problem to an optimization problem, minimizing Kullback-Leibler divergence between a simple proxy distribution and the posterior. This is equivalent to optimizing a lower bound to the marginal likelihood of the model, termed evidence lower bound (ELBO)~\citep{Jordan:1999ti}. In particular, we use stochastic variational inference (SVI)~\citep{JMLR:v14:hoffman13a}, which optimizes the ELBO using stochastic gradient descent. 

We first carry-out the approximate marginalization over $\beta$, lower-bounding the likelihood term. We then show how we can decompose the ELBO into a part which is equivalent to LDA, and into another part which contains the GP prior and therefore is more complex. We list all modified updates on the local and global parameters, with detailed calculations given in the supplementary material.
 
\paragraph{Approximate marginalization.}
We first marginalize over $\beta$ in the augmented joint distribution \eqref{eq:augmented_joint}.
Unfortunately, the marginal likelihood term cannot be computed in closed-form. We use Jensen's inequality to obtain a lower bound on the log likelihood,
\begin{align}
	\log p(&w_{dn} | z_{dn}=k, u, t_d)\\
	    &= \log \mathbb E_{p(\beta_{k\cdot t_d} | u)} \left[ p(w_{dn} | z_{dn}=k, \beta_{k\cdot t_d}\right]\nonumber\\
		&\ge \mathbb E_{p((\beta_{k\cdot t_d} | u)} \left[ \log \mathrm{Mult}\left(w_{dn} | \pi(\beta_{k\cdot t_d})\right) \right]\nonumber\\
		&=   K_{t_d \hat T} K_{\hat T \hat T}^{-1} u_{k} w_{dn}  - \mathbb E_{p((\beta_{k\cdot t_d} | u)} \left[\log \sum_w \exp(\beta_{kwt})  \right]\label{eq:expectation_intractable},
\end{align}
where $u_k$ is a $\hat T \times V$ matrix and $K_{t_d \hat T}$ is the $t_d$-th row of $K_{T \hat T}$.
The remaining expectation in \eqref{eq:expectation_intractable} is still intractable due to the sum inside of the logarithm. Following \citet{Blei:2006wj}, we introduce additional free variational parameters $\zeta_{kt}$ (see supplementary material). This results in a lower bound to $\log p(w_{dn} | z_{dn}=k, u, t_d)$:
\begin{align}
    \log &\tilde p(w_{dn} | z_{dn}=k, u, t_d) \label{eq:data_likelihood}\\
        &= K_{t_d \hat T} K_{\hat T \hat T}^{-1} u_{k} w_{dn}\nonumber\\
        &- \zeta_{k t_d}^{-1} \sum_w  \exp\left( K_{t_d \hat T} K_{\hat T \hat T}^{-1} u_{k w t_d} + \frac{\tilde K_{t_d t_d}}{2} \right)\nonumber\\
        &- \log(\zeta_{k t_d}) +1.\nonumber
\end{align}
Next, we use this lower bounded log-likelihood to derive a tractable variational objective which we can optimize.

\paragraph{Stochastic Variational Inference.}
We follow a variational structured mean-field approach \citep{Wainwright:2007du} and impose the following variational distributions on the latent variables,
$q(\theta_d|\lambda_d) = \text{Dir}(\lambda_d)$, $q(z_{dn}|\phi_{dn}) = \text{Mult}(\phi_{dn})$ and 
$q(u_{kw}|\mu_{kw}, \Sigma_{kw}) = \mathcal N_T(\mu_{kw}, \Sigma_{kw})$.
Eq. \eqref{eq:data_likelihood} gives rise to the following tractable lower bound of the marginal likelihood:
\begin{align}
	\mathcal L(\lambda, \phi, \mu, \Sigma, \zeta) = &\underbrace{\mathbb E_q[\log \tilde p(w | u,z)]}_{\mathcal L_1}\label{eq:ELBO_full}\\
	&+ \underbrace{\mathbb E_q[\log(p(z|\theta) p(\theta) p(u))]}_{{\mathcal L_2}}  + \underbrace{H(q)}_{\text{Entropy}}. \nonumber
\end{align}
The entropy term and $\mathcal L_2$ consist of standard results and a part that can be computed similarly as in standard LDA (see supplementary material). 
We also compute $\mathcal L_1$ in closed form:
\begin{align*}
	\mathcal L_1 = \sum_{t,k,w} &\sum_{d:t_d=t} n_{dw}\phi_{dwk}  \Biggl\{m_{kwt} - \log(\zeta_{kt}) + 1\\
    &\left.- \zeta_{kt}^{-1}  \sum_{w^\prime} \exp\left( m_{kw^\prime t}  + \frac{1}{2} (\Lambda_{kw^\prime t} + \tilde K_{t t}) \right)\right\},
\end{align*}
with
\begin{align*}
    m_{kwt} &= K_{t \hat T} K_{\hat T \hat T}^{-1}  \mu_{kw}\\
    \Lambda_{kwt} &=  K_{t \hat T} K_{\hat T \hat T}^{-1}  \Sigma_{kw}  K_{\hat T \hat T}^{-1} K_{\hat T t}.
\end{align*}
Objective $\mathcal L$ is optimized using SVI~\citep{JMLR:v14:hoffman13a}, i.e. for global variational parameters, we follow noisy natural gradients based on minibatches. Local variational parameter updates are similar to those in \citep{Wang2008continuous} and we do not replicate them here. Further details are provided in the supplementary material.


\paragraph{Global updates.}
We consider the Gaussian distributions $q(u_{kw})$ in \emph{natural parameterization}, i.e. using the parameters $\eta_{kw}^{(1)}= \Sigma^{-1}_{kw} \mu_{kw}$ and $\eta_{kw}^{(2)}= -\frac{1}{2}\Sigma^{-1}_{kw}$, where $\mu_{kw}$ are the Gaussian means and $\Sigma_{kw}$ the covariances. 
In SVI, we update these global parameters using stochastic estimates of the natural gradient and it turns out that in this case natural parameters result in simpler and more effective updates.

More specifically, for a Gaussian distribution, properties of the Fisher information matrix expose the simplification that the \emph{natural gradient} w.r.t. the natural parameters can be expressed in terms of the \emph{Euclidean gradient} w.r.t. the canonical parameters (i.e. mean and covariance). Namely, in general it holds for objectives $\cal F$ that depend on a Gaussian distribution that
\begin{align}
   \hat\nabla_{(\eta_1, \eta_2)} {\cal F}(\eta) = \big(\nabla_\mu{\cal F}(\eta) - 2\nabla_\Sigma{\cal F}(\eta)\mu,\; \nabla_\Sigma{\cal F}(\eta)\big), \label{eq:nat_grad_Gaussian}
\end{align}
where $\hat\nabla$ denotes the natural gradient and $\nabla$ the Euclidean gradient.
Applying \eqref{eq:nat_grad_Gaussian} to the variational objective \eqref{eq:ELBO_full}, we obtain
\begin{align}
    \begin{split} 
        \hat\nabla_{\eta_{kw}^{(1)}}\mathcal L &= \Xi_{kw}  +  B_{kw}\circ\left(m_{kw} - 1\right)  - \eta_{kw}^{(1)},\\
        \hat\nabla_{\eta_{kw}^{(2)}}\mathcal L &= -\frac{1}{2}K_{\hat T \hat T}^{-1} -\frac{1}{2}  C_{kw}  - \eta_{kw}^{(2)}.
    \end{split} \label{eq:natural_grad_Gaussian}
\end{align}
We used the following abbreviations: 
\begin{align*}
    \Xi_{kw} &= K_{\hat T \hat T}^{-1} \sum_{t}\sum_{d:t_d=t} K_{\hat T t} n_{dw}\phi_{dwk},\\
    B_{kw} &= \sum_{t}\sum_{d:t_d=t} \zeta_{k t}^{-1} n_{dw}\phi_{dwk}\\ &\qquad\qquad \times \exp\left(  m_{kwt} + \frac{ \Lambda_{kwt} +  \tilde K_{t t}}{2} \right)K_{\hat T \hat T}^{-1}K_{\hat T t},\\
    C_{kw} &= B_{kw}K_{ t \hat T} K_{\hat T \hat T}^{-1}.
\end{align*}
Above, $\circ$ denotes the Hadamard product. Details are provided the supplementary material.
Iterating through those updates completes the algorithm.


%% file: experiments.tex
\begin{figure*}[t]
  \centering
  \begin{minipage}[b]{0.32\textwidth}
    \includegraphics[width=\textwidth]{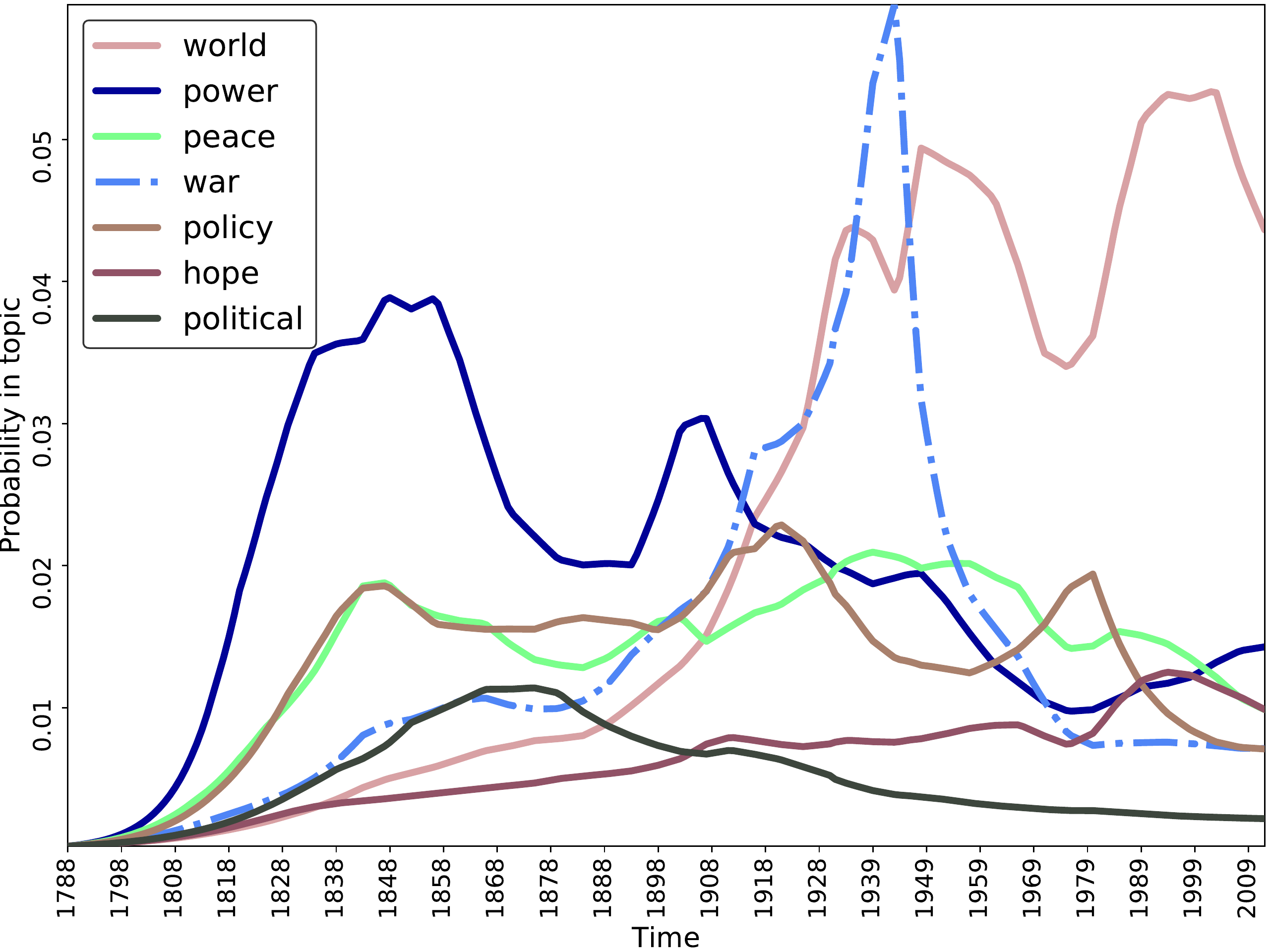}
    \label{fig:qual_11}
  \end{minipage}
  \hfill
  \begin{minipage}[b]{0.32\textwidth}
    \includegraphics[width=\textwidth]{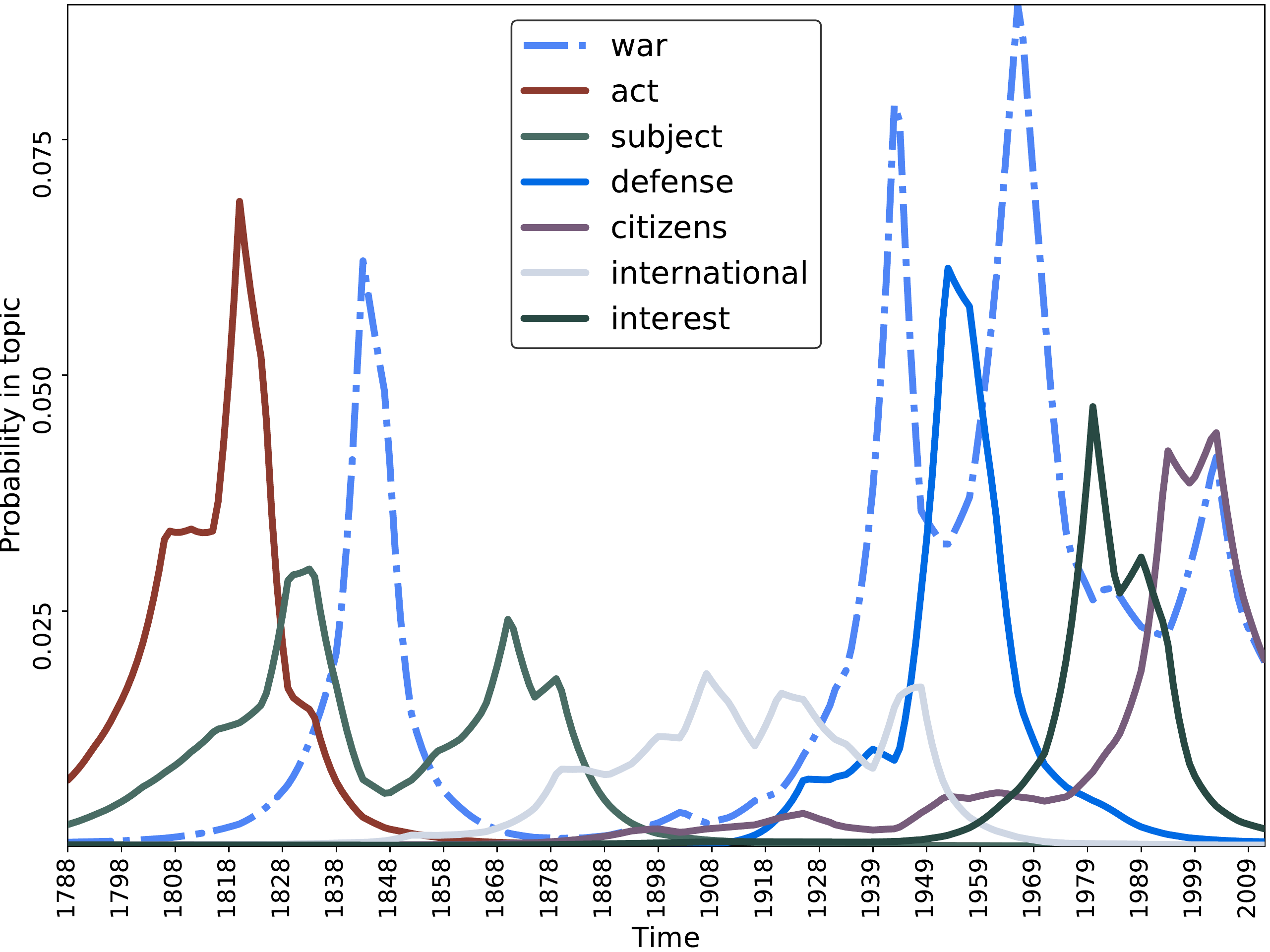}
    \label{fig:qual_12}
  \end{minipage}
   \hfill
  \begin{minipage}[b]{0.32\textwidth}
    \includegraphics[width=\textwidth]{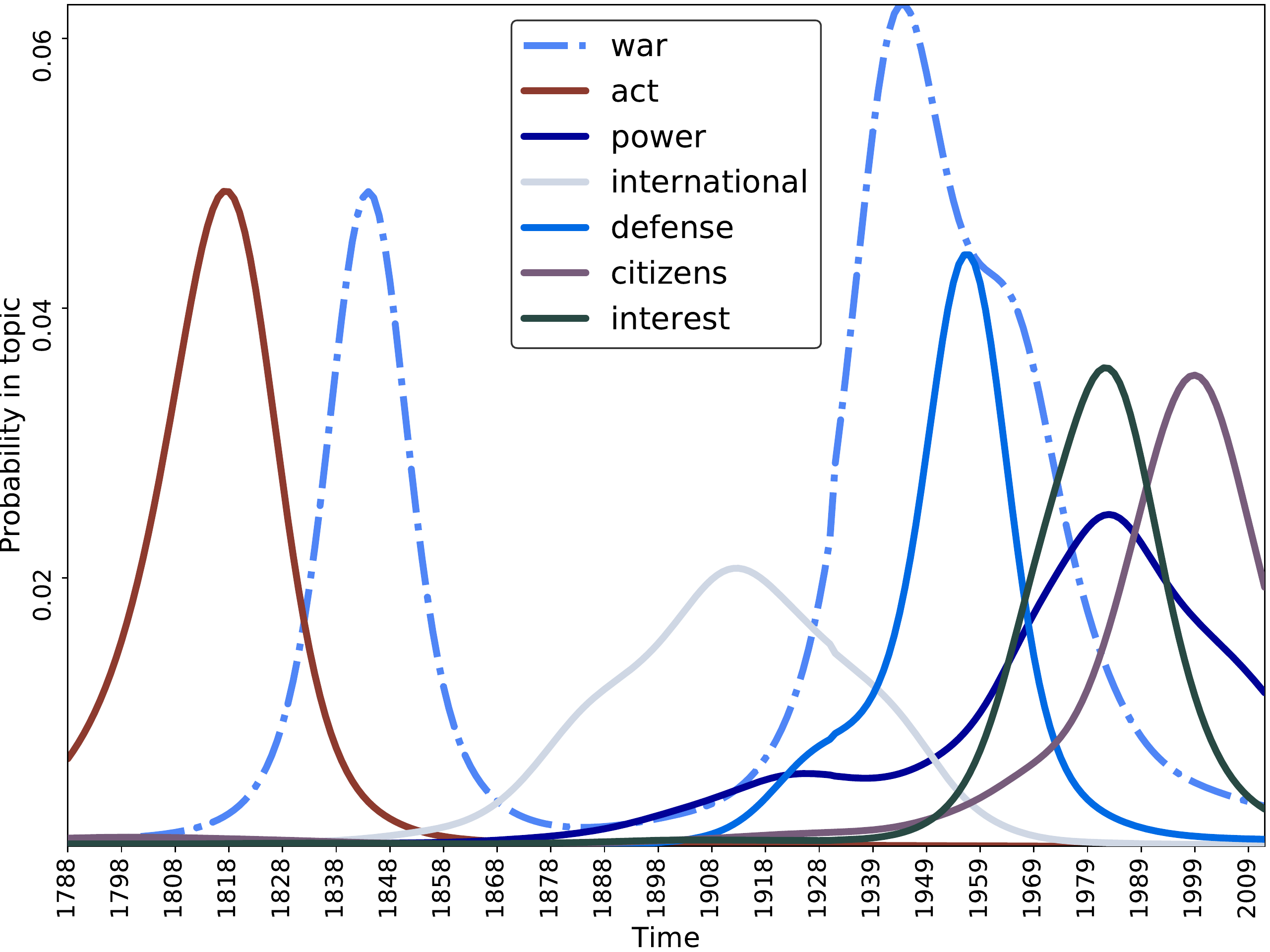}
    \label{fig:qual_13}
  \end{minipage}
  \caption{SoU: Learned word trajectories of the "war" topic using the Wiener kernel (left), OU kernel (middle) and Cauchy kernel (right). The Cauchy kernel provides smoother trajectories yet the OU kernel is able to provide a better resolution in time. Both outperform the baseline in terms of perplexity.}
  \label{fig:qual_war}
\end{figure*}
 \begin{figure*}[t]
  \centering
  \begin{minipage}[b]{0.32\textwidth}
    \includegraphics[width=\textwidth]{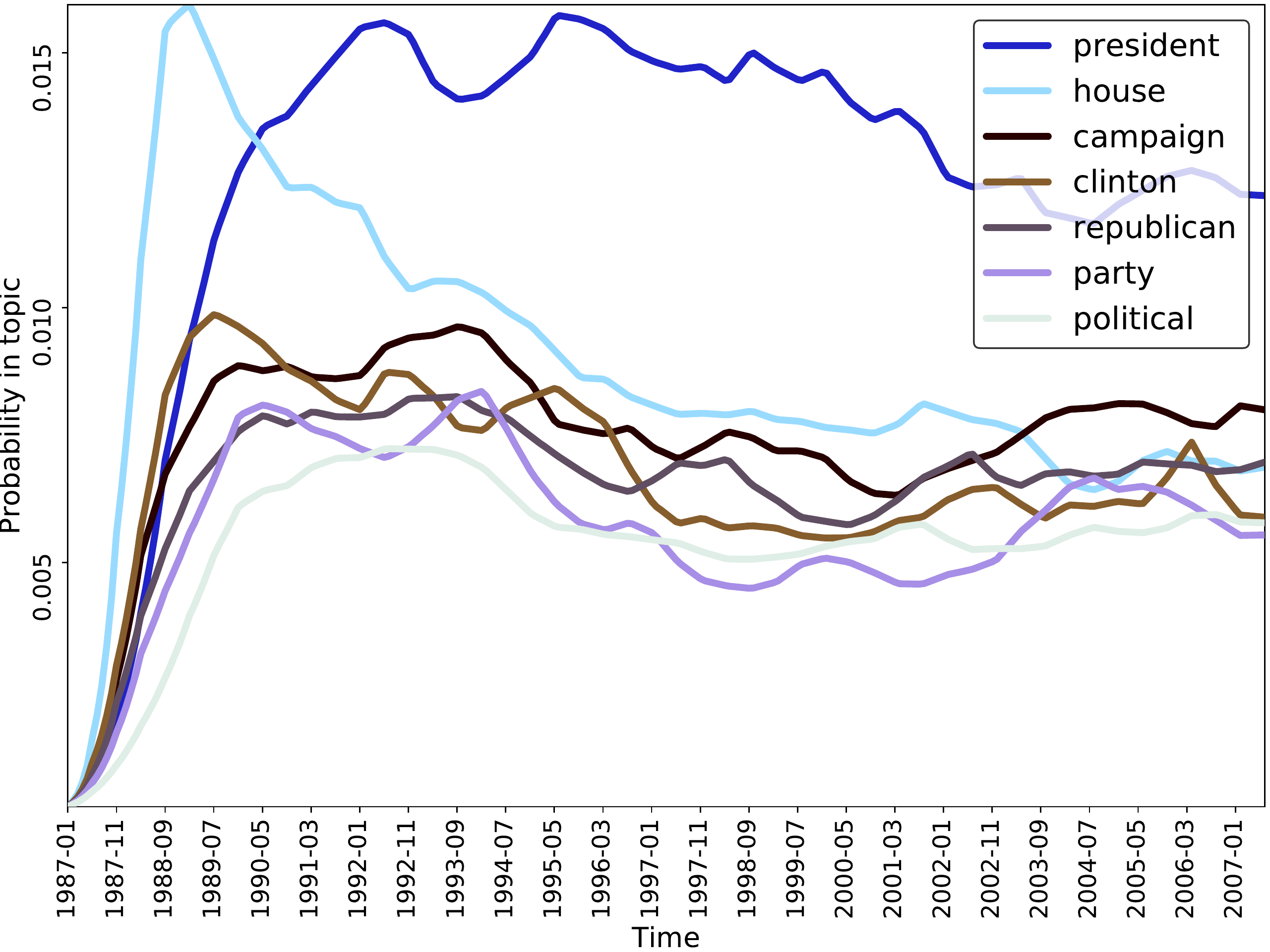}
    \label{fig:qual_21}
  \end{minipage}
  \hfill
  \begin{minipage}[b]{0.32\textwidth}
    \includegraphics[width=\textwidth]{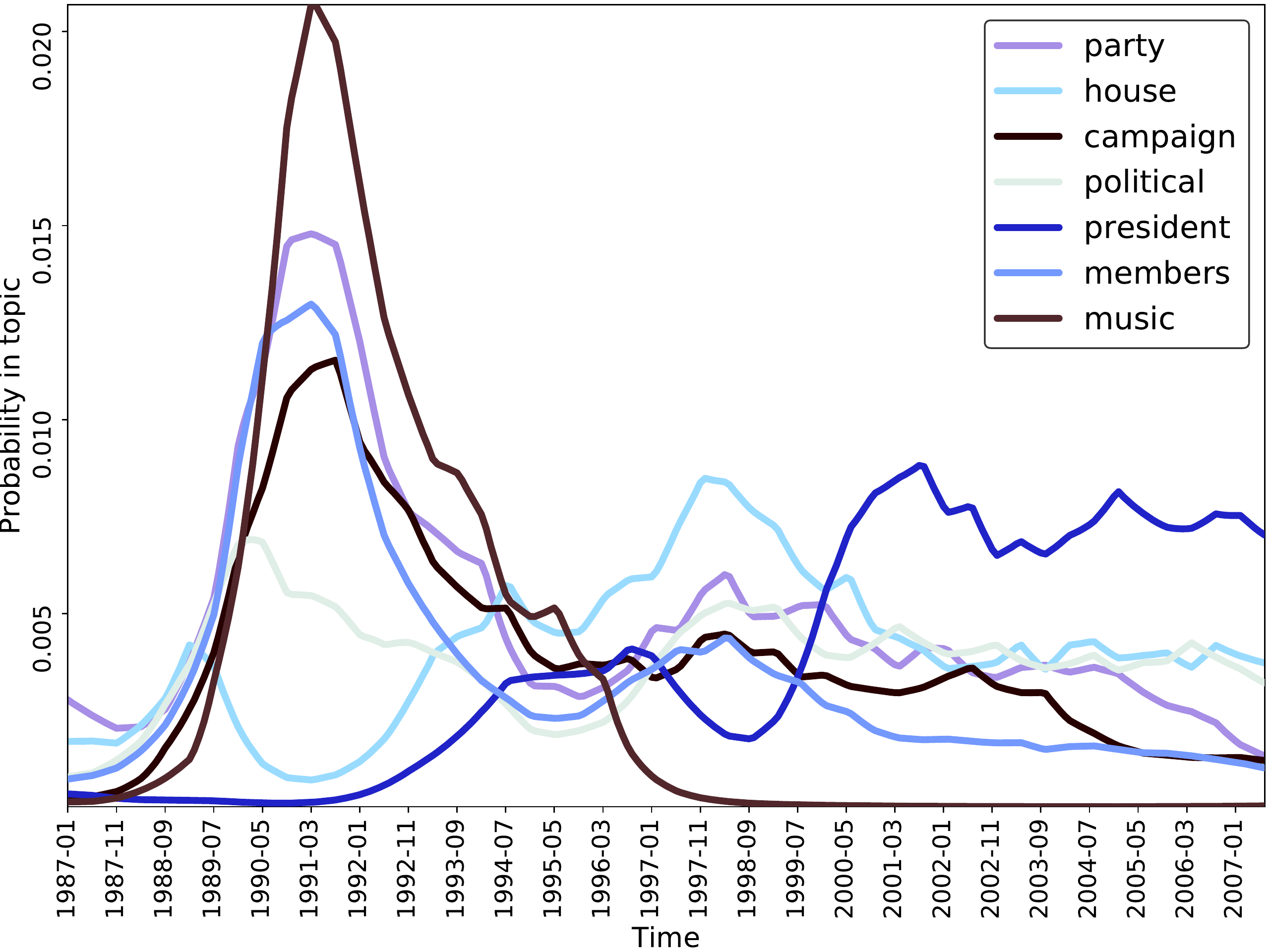}
    \label{fig:qual_22}
  \end{minipage}
   \hfill
  \begin{minipage}[b]{0.32\textwidth}
    \includegraphics[width=\textwidth]{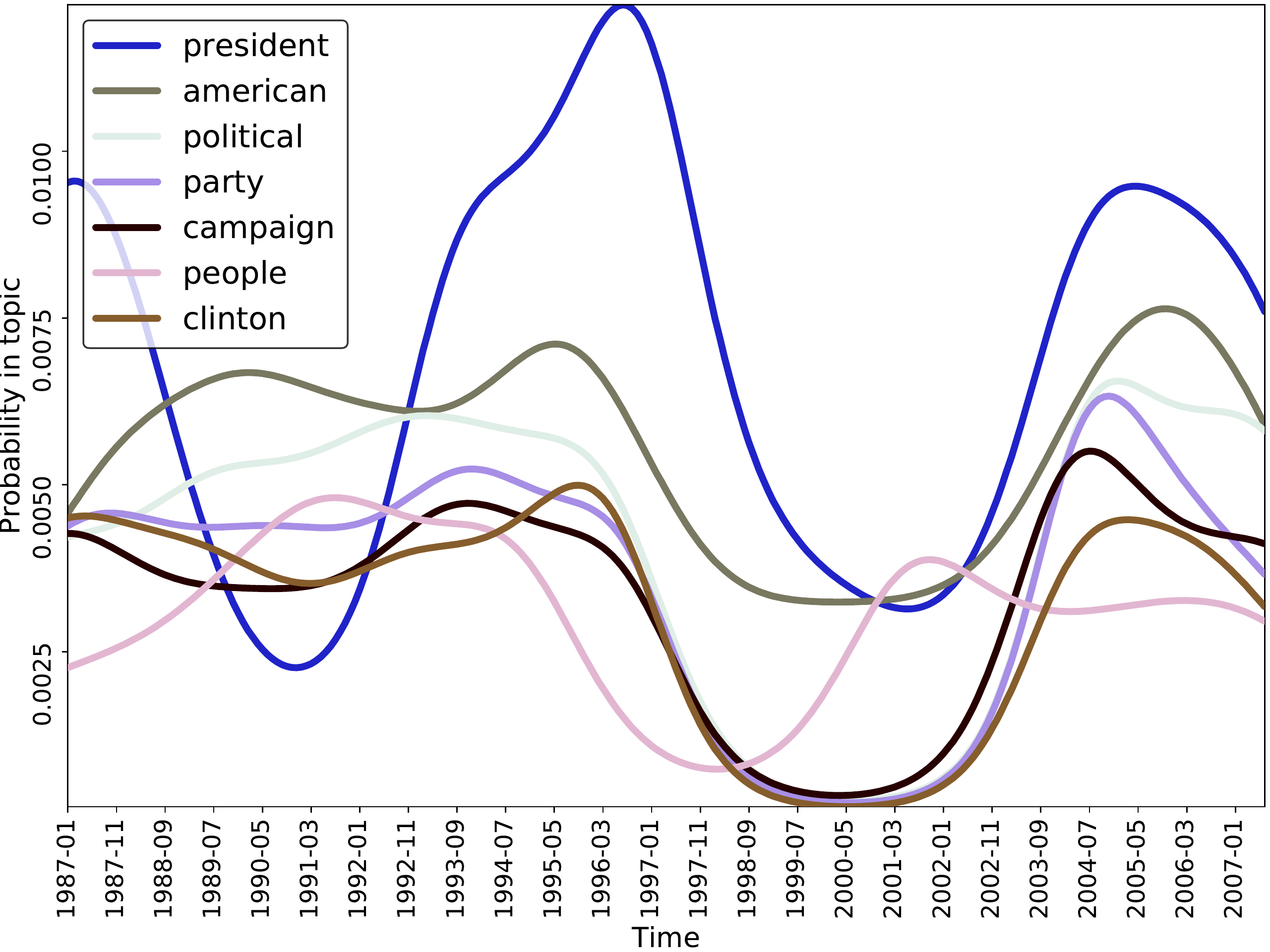}
    \label{fig:qual_23}
  \end{minipage}
  \caption{NYT: Learned word trajectories of the "election campaign" topic using the Wiener kernel (left), OU kernel (middle) and Cauchy kernel (right), which results in the smoothest curves.}
  \label{fig:qual_campaign}
\end{figure*}
 \begin{figure*}[t]
  \centering
  \begin{minipage}[b]{0.32\textwidth}
    \includegraphics[width=\textwidth]{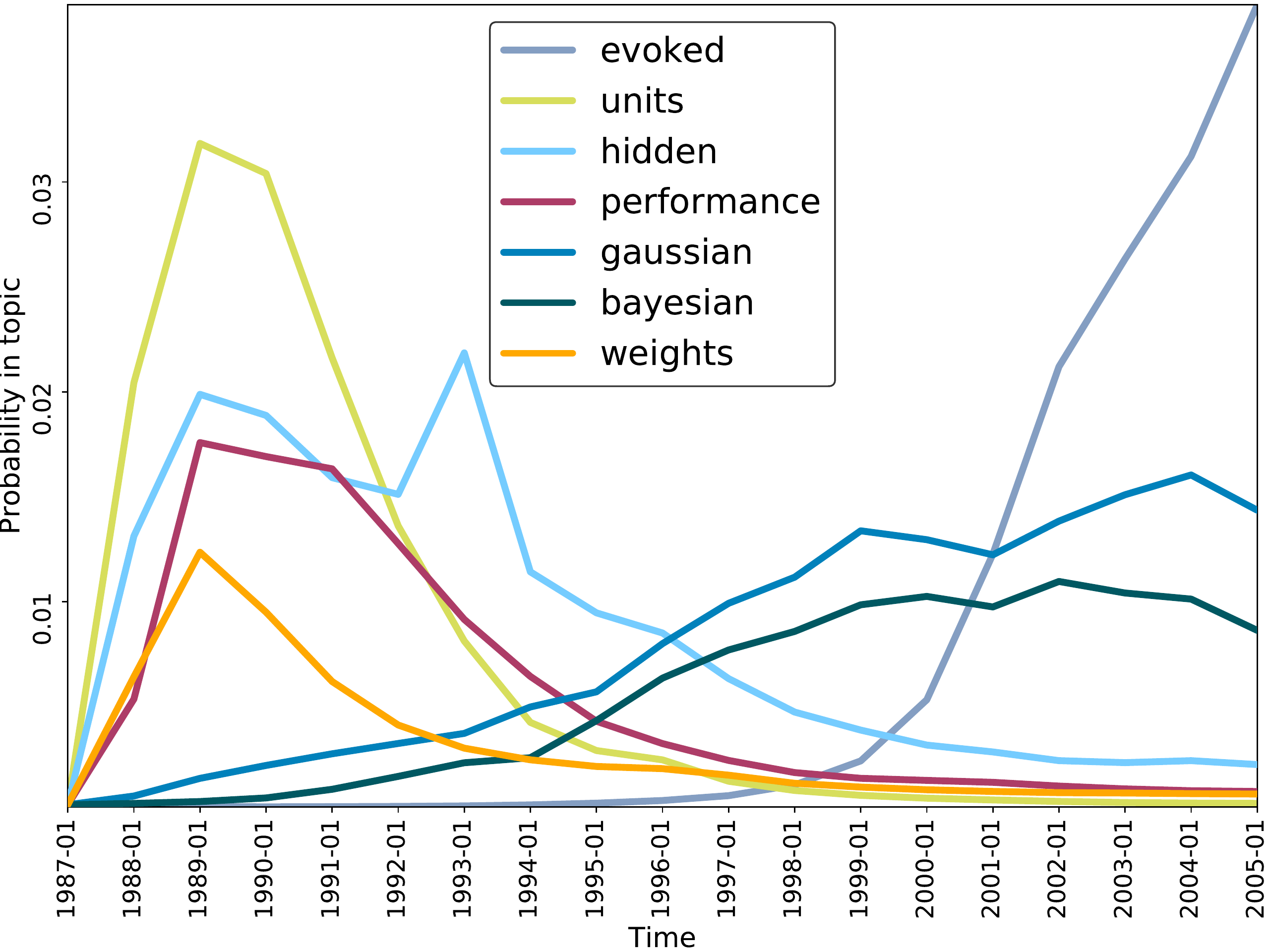}
    \label{fig:qual_31}
  \end{minipage}
  \hfill
  \begin{minipage}[b]{0.32\textwidth}
    \includegraphics[width=\textwidth]{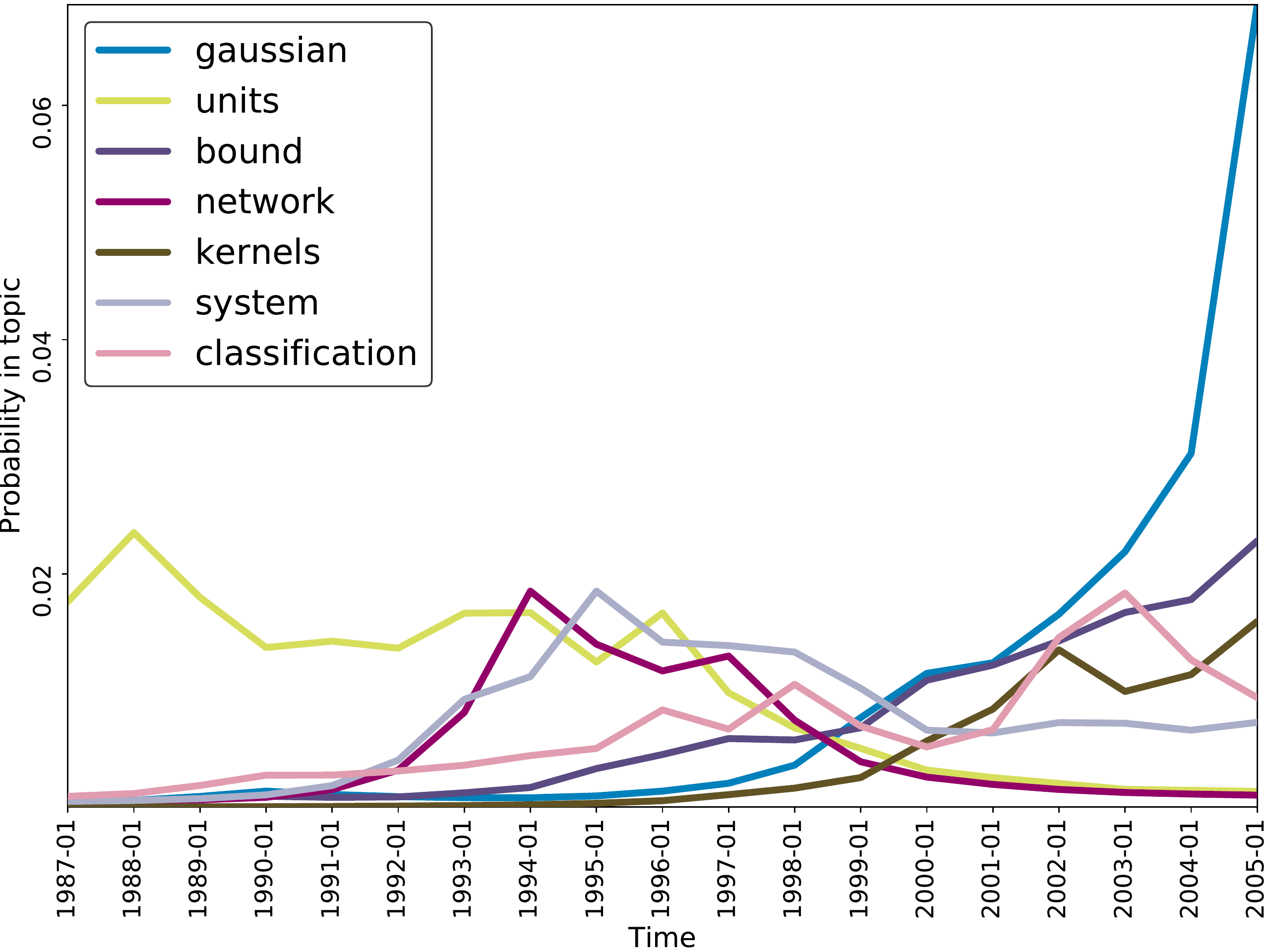}
    \label{fig:qual_32}
  \end{minipage}
   \hfill
  \begin{minipage}[b]{0.32\textwidth}
    \includegraphics[width=\textwidth]{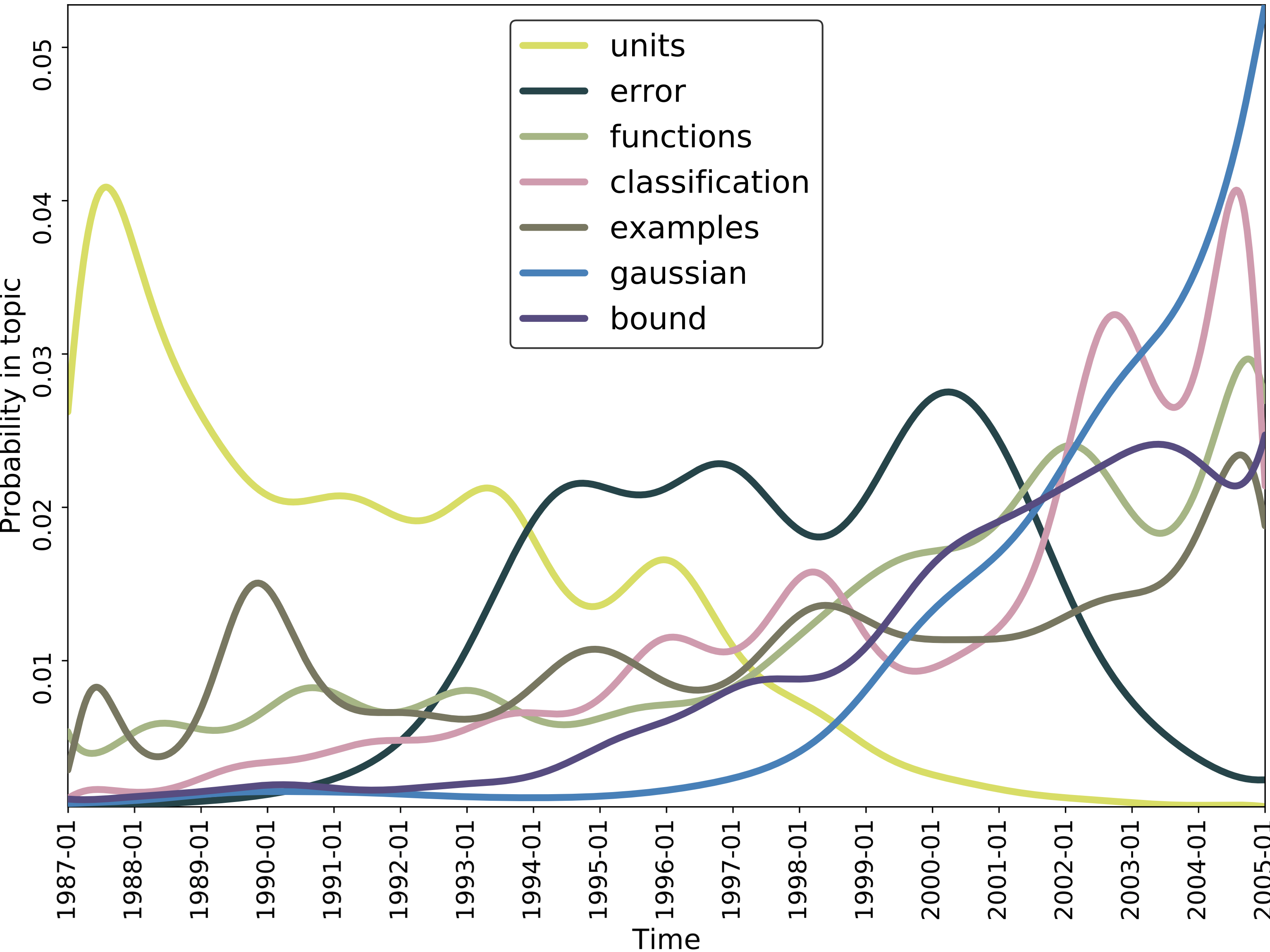}
    \label{fig:qual_33}
  \end{minipage}
  \caption{NIPS: Learned word trajectories of the "function approximation" topic using the Wiener kernel (left), OU kernel (middle) and Cauchy kernel (right). All three approaches identify terms that gain or loose importance within the topic over time. Since the Cauchy kernel shares statistical strength over a broader time horizon, its word trajectories are smoother.}
  \label{fig:qual_image}
\end{figure*}

\section{Experiments}
\label{sec:experiments}

We evaluate our method on three time-stamped text corpora.
Compared to standard DTMs with Wiener kernels, we find that incorporating other dynamic priors may lead to improved predictive likelihoods and perplexity on held-out data.
Using different kernel functions within our framework, we find new insights in the data that could not be found using the classic DTM of \citet{Wang2008continuous}, which uses the Wiener kernel 
and thus results in an unbounded variance over the time span. 
This promotes topics that are consistent, albeit relatively static. 

By making use of the greater flexibility that comes with general GPs, we show how to extend and enhance an analysis.
For instance, by using the \emph{OU kernel}, we introduce a mean reverting force that quickly "draws" word probabilities towards zero, resulting in topics that are consistent and constrained in time and more sensitive to changes.
Further, in situations in which the classic approach collapses 
most of the probability mass to single words per time stamp, we compensate by using \emph{Cauchy kernels}, which place a smooth filter on word probabilities onto the topic over time.
On the other hand, more fine-grained temporal dynamics can be captured by \emph{RBF kernels}, due to their short-range memory.

\paragraph{Data and preprocessing.}\mbox{}\\
1. We use the ``The New York Times Annotated Corpus'' (\textbf{NYT}) \citep{TheNewYorkTimesA:2008vx}, which consists of over 1.8 million articles published between 1987 and 2007 with $T = 7475$ unique time stamps. We subsample 100000 documents.

2. We use the \textbf{NIPS} dataset that contains 2711 papers from the NIPS conferences between 1987 and 2006\footnote{\url{http://www.datalab.uci.edu/author-topic/NIPs.htm}} resulting in $T = 19$ time stamps.

3. We use the ``State of the Union'' (\textbf{SoU}) addresses of U.S. presidents, which span more than two centuries, resulting in $T = 224$ different time stamps \footnote{\url{http://www.presidency.ucsb.edu/sou.php}}. We increase the number of documents to $4428$ by treating every chunk of ten paragraphs in a speech as a separate document.

For preprocessing, we filtered the raw data using a standard stop word list. After collecting word statistics, we remove words that appear less than 25 times across a whole corpus. We further shrink the vocabulary by removing words whose score is less than a certain threshold, resulting in dictionaries of reasonable size (see supplementary material). After this step, we remove documents with effective lengths less than ten word occurrences.
We initialize our models by randomly selecting $K$ documents for any given time stamp and setting probabilities in topic $k$ of occurring words proportional to their frequencies in the $k$-th document.

\paragraph{Hyperparameters.}
In our experiments we select the hyperparameters via grid search but they could also be directly learned in our inference scheme using the approximate empirical Bayes approach \citep{EB89}. 

\paragraph{Qualitative Results.}
We now discuss the qualitative results obtained from applying our model on all three corpora. For certain example topics, we plot and discuss the probabilities of the most important words in this topic over time. As a general tendency, we find that our proposed Ornstein-Uhlenbeck and Cauchy kernels outperform the standard Wiener kernel in terms of interpretability and in terms of usefulness for detecting events.

\emph{SoU.} We consider a topic of with war and peace, found when fitting our generalized DTM to the state-of-the-union corpus. Figure~\ref{fig:qual_war} shows the word probabilities within this topic over time for all three considered kernels.
The Wiener kernel is able to find a semantically coherent word distribution for this topic. We observe a relatively high probability of the term \emph{"war"} over the whole time span with a sharp peak around 1939 (World War II). Using the Cauchy kernel, we are able to gain a better resolution of the dynamics for the importance of this term. We observe two separate high-probability periods of the word "war". One is matching the time of the American-Mexican war 1846-1848, the other one the World Wars and Vietnam war. We attribute this finding to the fact that the Cauchy kernel shares more statistical strength over time due to its long-term memory property.

While this model already provides a better insight into active time periods of the topic, additionally introducing a mean-reverting force via the OU kernel provides a mean to "super-resolve" topic activity quite accurately to certain events. We observe high probability for the term \emph{"war"} again around 1848, a small plateau in the 1910s (World War I) rising to a high value in 1939 (World War II) and the 1960s (Vietnam war). We even observe a small bump in the beginning and through the 1980s (possibly the war in Afghanistan) and another peak in the mid 2000s (second Afghanistan war). 
Additionally, when looking at the words with highest probability at these times, we observe that the model is able to place probability mass on terms relating to the different wars, e.g. \emph{"texas"} for the American-Mexican war (which was fought over Texas) or \emph{"attack"} and \emph{"japanese"} in 1942 (where the attack on Pearl Harbor took place). Based on these findings, the Ornstein-Uhlenbeck kernel seemed most appropriate for this task.

\emph{NYT.} Another interesting use case scenario is the analysis of news texts. One of the topics identified when analyzing the New York Times corpus deals with presidential election campaigns.
Figure~\ref{fig:qual_campaign} shows probability trajectories of terms in that topic for the Wiener (left), Ornstein-Uhlenbeck (middle), and Cauchy (right) kernels, respectively. 
We observe, that the baseline model (Wiener kernel) is able to capture meaningful terms. Going beyond this, the OU kernel arguably reflects the election campaigns in 1992 and 1996. The Cauchy kernel results in even smoother trajectories. 
These findings, however, deserve a more thorough investigation and interpretation. Nevertheless, this example shows that different kernels reveal qualitatively different phenomena.

\emph{NIPS.} Applying generalized DTMs on the NIPS corpus allows us to track trends in machine learning over the last 20 years. We present probability trajectories of a topic related to classification and function approximation. Again, we show results for Wiener, OU, and Cauchy kernels (Figure~\ref{fig:qual_image}, left to right). We observe from the baseline model that neural networks gained attention in the late 1980s and early 1990s. However, excitement subsided in the late 1990s and Bayesian methods were on the rise (our data set is not recent enough to detect the uptrend of neural networks in the last 5-10 years). While the Wiener kernel models overall development, the Ornstein-Uhlenbeck process is able to better react to small scale changes, resulting in a more realistic representation of term development. Additionally, it finds more general terms, such as \emph{"network"}, \emph{"classification"} and \emph{"system"}. Using the Cauchy kernel with its long-term memory prevents from placing large probability mass on the rapidly rising term \emph{"gaussian"}. 
The Cauchy kernel is also able to identify more general terms.
\paragraph{Quantitative Results.}
We show that using our approach not only leads to interesting dynamic topics but also generalizes better to unseen data. We use all documents associated with time stamps $T_{\text{train}}$ as training set and analyze the predictive held-out likelihoods on remaining documents (associated with the time stamps $T_{\text{test}} = T \setminus T_{\text{train}}$).
We experiment on the NYT dataset and randomly select $T_{\text{train}}$ to hold 85\% of the unique time stamps.
    
  \begin{figure}[!h]
    \centering
    \includegraphics[width=.48\textwidth]{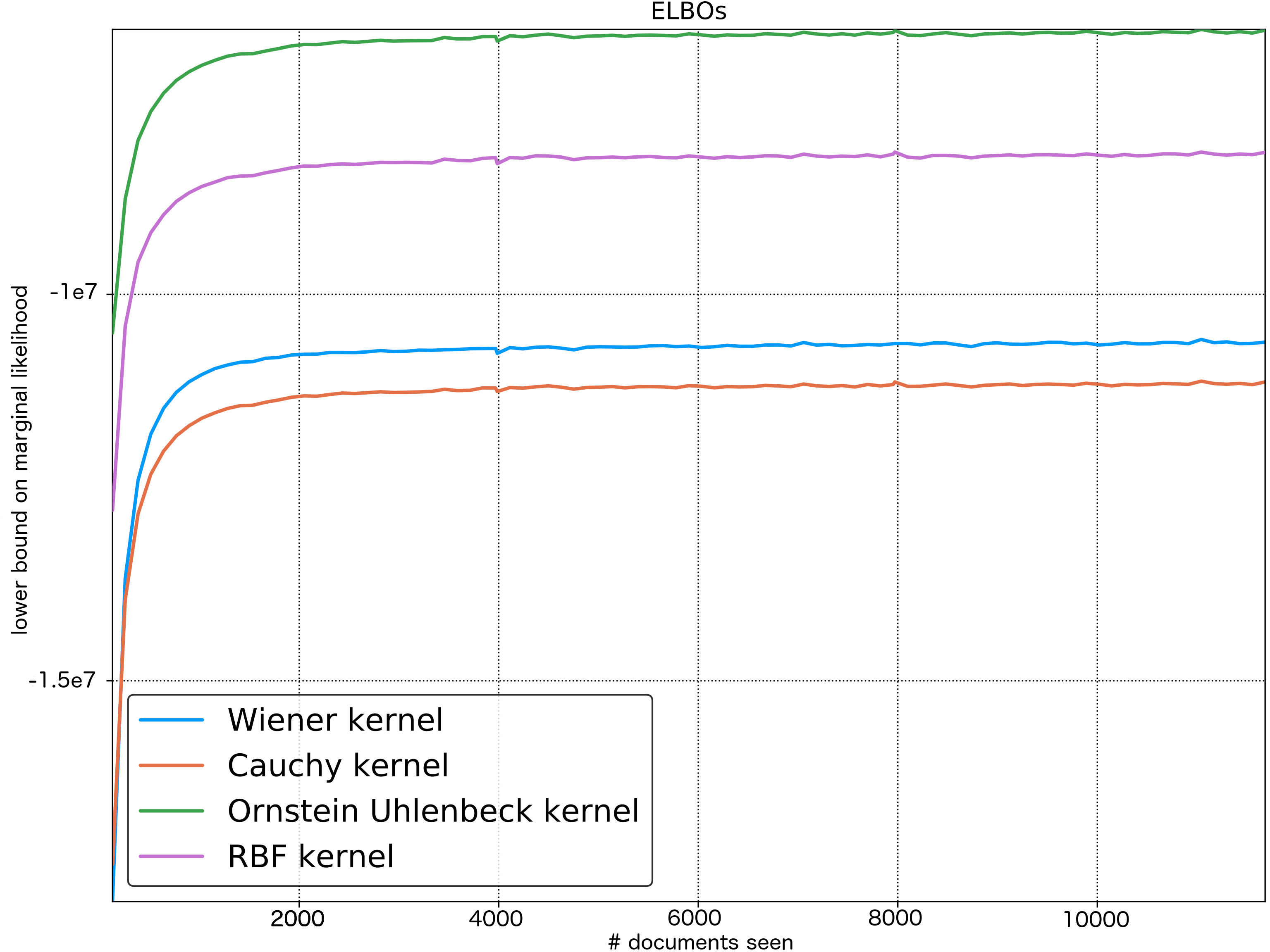}

    \caption{SoU: Evidence lower bound against the number of documents seen. On all used kernels, the objective function converges to an optimum.} \label{fig:pred_likelihoods_nyt}
  \end{figure}
    \begin{table}[!h]
    \centering
        \footnotesize
        \begin{tabular}{l|l|l|l|l}
         Data & cDTM & gDTM& gDTM& gDTM\\
                & (baseline) & OU & Cauchy & RBF\\
        \hline
        NYT & 1.42323 & \textbf{1.42073} & 1.42129 & 1.42374 \\
        \hline
        NIPS & 1.4931 & 1.48149 & \textbf{1.48105} & 1.4821 \\  
        \hline
        SoU & 1.46854 & 1.45594 & \textbf{1.45575} & 1.46023 \\
        \hline
        \end{tabular}
      \caption{Per-word predictive perplexities (lower numbers are better). We constantly outperform the baseline on all data sets.} \label{tab:perplexity}
      
    \end{table}

\normalsize
\raggedright
Table~\ref{tab:perplexity} shows that our method outperforms the baseline in terms of per-word predictive perplexity~\citep[e.g.][]{Blei:2007vt}. We observe that the perplexity on both the NIPS and SoU dataset is best when the dynamics are modeled by a GP with Cauchy kernel while the NYT dataset is best captured by a OU kernel. This shows again the advantage of using our approach over the state-of-the-art. Having the flexibility of modeling the dynamics by a GP we can account for the different dynamics that may underlie different datasets. Additionally, Figure~\ref{fig:pred_likelihoods_nyt} shows the ELBO objective function when fitting a model to the SoU data set, eventually reaching an optimum. Results on the different data sets were similar.

\paragraph{Remarks.}
We argue that as common in probabilistic modeling, the prior should not be chosen based on predictive likelihood alone. Instead, a prior is a modeling choice that helps reveal the effects that one searches for. 
Depending on the problem at hand, a practitioner would  choose the suitable kernel, be it the Wiener kernel, Ornstein-Uhenbeck kernel, RBF kernel or Cauchy kernel.
The Ornstein-Uhlenbeck kernel has the favorable property of localizing topics in time, which may be a promising tool for event detection. However, if the length scale is too small, topics change their word distributions at a frequency which is too high, in which case the results are less interpretable. On the other hand, the RBF kernel (and even more so the Cauchy kernel) has long-time memory and is generally more data efficient, which has advantages if the data set is small. Ultimately, many other kernels may be designed for different purposes.


%% file: conclusion.tex
\section{Conclusion and Future Work} 
\label{sec:conclusion_and_future_work}

We presented the generalized dynamic topic model, which allows for dynamic topic modeling with a broader class of dynamic priors, and which easily scales up to very large text collections. In particular, we generalized dynamic topic models from Brownian motion priors to arbitrary Gaussian process priors. We showed in our experiments that our approach leads to better predictive likelihoods on held-out documents, and to interesting new qualitative findings, such as temporally localized topics, and topics that display long-range temporal dependencies. As a possible future extensions, we plan to consider periodic kernels for repeating events, and to extend dynamic topic modeling from the time domain to the geo-spatial domain, such as text equipped with location information.

%% file: appendix.tex
\onecolumn
\section{Appendix}
\subsection{Approximate marginalization}
\label{app:marginalize}
Following \citep{Blei:2009uo}, we lower bound the intractable expectation in \eqref{eq:expectation_intractable} by computing the first order Taylor approximation of the logarithm around an arbitrary location parameter $\zeta_{kt}>0$,

\begin{align*} 
\mathbb E_{p(\beta_t | u)} &\left[\log \sum_v \exp(\beta_{k v t})  \right] \le \zeta_{k t}^{-1} \sum_v  \exp\left( K_{t \hat T} K_{\hat T \hat T}^{-1} u_{k v .} + \frac{\tilde K_{t t}}{2} \right) + \log(\zeta_{k t}) - 1.
\end{align*}

\paragraph{Updating the Taylor expansion location parameters.}
In each iteration in our inference algorithm we optimize the location parameter of the Taylor expansion to achieve the tightest possible bound on true marginal likelihood (c.f. equation \ref{eq:data_likelihood}).
Setting the derivative of $\cal L$ w.r.t. $\zeta_{kt}$ to zero and solving for $\zeta_{kt}$ gives the update

\begin{align*}
	\zeta_{kt} = \sum_v \exp\left( m_{kv t}  + \frac{1}{2} (\Lambda_{kvt} + \tilde K_{t t}) \right).
\end{align*}

\subsection{Derivation of the Variational Objective}
\label{app:var_objective}
Recall the variational objective
\begin{align*}
    \mathcal L(\lambda, \phi, \mu, \Sigma) = \mathbb E_q[\log \tilde p(w | u,z) p(z|\theta) p(\theta) p(u)] - \mathbb E_q[\log q(\theta)q(z)q(u)].
\end{align*}
The first term is
\begin{align*}
    &\EE_q\left[ \log \tilde p(w| z,u) \right]\\  
    &= \sum_{t,n,k} \EE_q\left[ z_{tnk} \log \tilde p(w_{tn} | z_{tn}=k, u)\right] \\
    &= \sum_{t,n,k} \EE_q[z_{tnk}]  \left\{ K_{t \hat T} K_{\hat T \hat T}^{-1} \EE_q[u_{k . .}] w_{tn}  - \zeta_{k t}^{-1} \sum_v  \EE_q\left[\exp\left( K_{t \hat T} K_{\hat T \hat T}^{-1} u_{k v .}  + \frac{\tilde K_{t t}}{2} \right)\right] - \log(\zeta_{k t}) + 1\right\}\\
    &= \sum_{t,n,k} \phi_{tnk}  \Big\{  K_{t \hat T} K_{\hat T \hat T}^{-1} \mu_{k . .} w_{tn}  - \zeta_{k t}^{-1}  \sum_v \exp\left( K_{t \hat T} K_{\hat T \hat T}^{-1} \mu_{k v}  + \frac{1}{2} (K_{t \hat T} K_{\hat T \hat T}^{-1} \Sigma_{kv}  K_{\hat T \hat T}^{-1} K_{\hat T t} + \tilde K_{t t}) \right)\\
    &\quad - \log(\zeta_{k t}) + 1\Big\}\\\
     &= \sum_{t,n,k} \phi_{tnk}  \Big\{w_{tn.}^\top m_{k. t}  - \zeta_{k t}^{-1}  \sum_v \exp\left( m_{kv t}  + \frac{1}{2} (\Lambda_{kvt} + \tilde K_{t t}) \right)- \log(\zeta_{k t}) + 1\Big\},
\end{align*}
where $m_{kv t} = K_{t \hat T} K_{\hat T \hat T}^{-1}  \mu_{kv}$ and $\Lambda_{kvt} =  K_{t \hat T} K_{\hat T \hat T}^{-1}  \Sigma_{kv}  K_{\hat T \hat T}^{-1} K_{\hat T t}$.

The second term is

\begin{align*}
    \EE_q[\log p(z |\theta)] &= \sum_{t,n} \EE_q[\log p(z_{tn}|\theta_t)] = \sum_{t,n,k} \phi_{tnk} \EE_q[\log \theta_{tk}]\\
    &= \sum_{t,n,k} \phi_{tnk} \left(\psi(\lambda_{t k}) - \psi\left( \lambda_{t 0} \right)\right),
\end{align*}
where $\lambda_{t0} = \sum_{k}\lambda_{d k}$.

The negative KL terms are

\begin{align*}
    \EE_q[\log p(u ) - \log q(u)] = -\sum_{k,v}\mathrm{KL}(q(u_{kv})||p(u_{kv})) &\uptoconst -\frac{1}{2}\sum_{k,v} \left( \mu_{kv} K^{-1}_{\hat T \hat T} \mu_{kv} + \tr(\Sigma_{kv} K^{-1}_{\hat T \hat T}) - \log |\Sigma_{kv}|\right).
\end{align*}
and

\begin{align*}
    \EE_q[\log p(\theta ) - \log q(\theta)] &= -\sum_{t}\mathrm{KL}(q(\theta_t)||p(\theta_t))\\
    &\uptoconst \sum_{t,k} \left( (\alpha_k - \lambda_{tk}) (\psi(\lambda_{t k}) - \psi\left(\lambda_{t0}\right))  + \log \Gamma(\lambda_{tk})   \right)  - \Gamma(\lambda_{t0}).
\end{align*}
The entropy of $q(z)$ is

\begin{align*}
    -\EE_q[q(z)] = -\sum_{t,n,k} \phi_{tnk} \log \phi_{tnk}.
\end{align*}
Finally, summing all terms gives the variational objective

\begin{align*}
    \mathcal L(\lambda, \phi, \mu, \Sigma)
    &= \sum_{t,n,k} \phi_{dnk}  \Big\{w_{tn}^\top m_{k. t}  - \zeta_{k t}^{-1}  \sum_v \exp\left( m_{kv t}  + \frac{1}{2} (\Lambda_{kvt} + \tilde K_{t t}) \right)- \log(\zeta_{k t}) + 1\\
    &\quad+ \psi(\lambda_{t k}) - \psi\left( \lambda_{t 0} \right)     - \log \phi_{tnk}    \Big\}
    -\frac{1}{2}\sum_{k,v} \left( \mu_{kv} K^{-1}_{\hat T \hat T} \mu_{kv} + \tr(\Sigma_{kv} K^{-1}_{\hat T \hat T}) - \log |\Sigma_{kv}|\right)\\
    &\quad+ \sum_{t,k} \left( (\alpha_k - \lambda_{tk}) (\psi(\lambda_{t k}) - \psi\left(\lambda_{t0}\right))  + \log \Gamma(\lambda_{tk})   \right) - \Gamma(\lambda_{t0}) + \mathrm{const}.
\end{align*}


\subsection{SVI Updates}
\label{app:updates}
In the following we provide more details on how the the parameter updates are derived.
\paragraph{Updating the Taylor expansion location parameter}
The derivative of the variational objective with respect to the location parameter of the Taylor expansion is

\begin{align*}
		\frac{\partial {\mathcal  L}}{\partial \zeta_{kt}}	&= \sum_{n}  \phi_{tnk} \zeta_{kt}^{-1}\left( \zeta_{k t}^{-1} \sum_v \exp\left( m_{kv t}  + \frac{1}{2} (\Lambda_{kvt} + \tilde K_{t t}) \right) - 1\right).
\end{align*}
Stetting the derivative zero and solving for $\zeta_{kt}$ gives the update

\begin{align*}
	\zeta_{kt} = \sum_v \exp\left( m_{kv t}  + \frac{1}{2} (\Lambda_{kvt} + \tilde K_{t t}) \right).
\end{align*}

\paragraph{Updating the local variables of $q(z)$}
The derivative of $\cal L$ w.r.t. $\phi_{tnk}$ is

\begin{align*}
	\frac{\partial {\mathcal  L}}{\partial \phi_{tnk}} &= w_{tn}^\top m_{k. t}  - \zeta_{k t}^{-1}  \sum_v \exp\left( m_{kv t}  + \frac{1}{2} (\Lambda_{kvt} + \tilde K_{t t}) \right)- \log(\zeta_{k t}) + \psi(\lambda_{t k}) - \psi\left( \lambda_{t 0} \right)     - \log \phi_{tnk}
\end{align*} 
Setting the derivative zero leads to

\begin{align*}
	\phi_{tnk} &= \exp\big\{w_{tn}^\top m_{k. t}  - \zeta_{k t}^{-1}  \sum_v \exp\left( m_{kv t}  + \frac{1}{2} (\Lambda_{kvt} + \tilde K_{t t}) \right)- \log(\zeta_{k t}) + \psi(\lambda_{t k}) - \psi\left( \lambda_{t 0} \right)      \big\}.
\end{align*} 
Inserting the update of the previous update of $\zeta_{kt}$ this simplifies to
\begin{align*}
	\phi_{tnk} &= \exp\big\{w_{tn}^\top m_{k. t}  - 1 - \log(\zeta_{k t}) + \psi(\lambda_{t k}) - \psi\left( \lambda_{t 0}
	\right)      \big\}\\
	&\propto
	\exp\big\{w_{tn}^\top m_{k. t} - \log(\zeta_{k t}) + \psi(\lambda_{t k}) - \psi\left( \lambda_{t 0}\right)\big\}.
\end{align*}

The update for the parameter vector $\phi_{tn.}$ is obtained by renormalizing (such that $||\phi_{tn.}||_1 = 1$). 

\paragraph{Updating the global variables}
The standard Euclidean gradient of $\cal L$ with respect to the mean and covariance parameters of $q(u_{kv})$ is

\begin{align*}
    \frac{\partial {\mathcal  L}}{\partial \mu_{kv}} 
    &= \Xi_{kv}  - B_{kv} -K_{\hat T \hat T}^{-1}\mu_{kv},\\
 	\frac{\partial {\mathcal  L}}{\partial \Sigma_{kv}}
 	&= -\frac{1}{2} C_{kv} + \frac{1}{2}\Sigma_{kv}^{-1} - \frac{1}{2}K_{\hat T \hat T}^{-1},
\end{align*}
where $\Xi_{kv} = \sum_{t,n} \phi_{tnk} w_{tnv} K_{\hat T \hat T}^{-1}K_{\hat T t}$,
$B_{kv} = \sum_{t,n} \zeta_{k t}^{-1} \phi_{tnk}  \exp\left( m_{kvt} + \frac{\Lambda_{kvt} + \tilde K_{t t}}{2} \right)K_{\hat T \hat T}^{-1}K_{\hat T t}$ and
$C_{kv} = \sum_{t,n} \zeta_{k t}^{-1} \phi_{tnk}  \exp\left( m_{kvt} + \frac{\Lambda_{kvt} + \tilde K_{t t}}{2} \right)K_{\hat T \hat T}^{-1}K_{\hat T t}K_{ t \hat T} K_{\hat T \hat T}^{-1}$.

We now consider the Gaussian distributions $q(u_{kv})$ in natural parametrization using $\eta_{kv}^{(1)}= S^{-1}_{kv} \mu_{kv}$ and $\eta_{kv}^{(2)}= -\frac{1}{2}S^{-1}_{kv}$. 
Applying formula \ref{eq:nat_grad_Gaussian} we obtain the natural gradient w.r.t. natural parameters,

\begin{align*}
        \hat\nabla_{\eta_{kv}^{(1)}}\mathcal L &= \Xi_{kv}  - B_{kv} -K_{\hat T \hat T}^{-1}\mu_{kv} - 2(-\frac{1}{2} C_{kv} + \frac{1}{2}\Sigma_{kv}^{-1} - \frac{1}{2}K_{\hat T \hat T}^{-1})\mu_{kv} \\
            &= \Xi_{kv}  +  B_{kv}\circ\left(  m_{kv} - 1\right)  - \eta_{kv}^{(1)}
\end{align*}
and
\begin{align*}
        \hat\nabla_{\eta_{kv}^{(2)}}\mathcal L &= -\frac{1}{2}  C_{kv} - \frac{1}{2}K_{\hat T \hat T}^{-1} - \eta_{kv}^{(2)}.           
\end{align*}

Note that $m_{kv}$ as function of the natural parameters is
\begin{align*}
        m_{kv} &= K_{T \hat T} K_{\hat T \hat T}^{-1} \mu_{kv}
        = -\frac{1}{2}K_{T \hat T} K_{\hat T \hat T}^{-1}\left(\eta_{kv}^{(2)}\right)^{-1}\eta_{kv}^{(1)}.
\end{align*}

\subsection{Global td-idf score}
\label{app:tfidf}
To determine important words, we use an extension to the classic tf-idf scoring scheme.
The score of a word $$\text{score}(w) = \frac{n_w}{M}\ln\left(\frac{D}{n_{dw}}\right)$$ where $M$ is the total amount of terms in the corpus, $D$ is the number of documents, $n_{dw}$ is the frequency of word $w$ in document $d$ and $n_w = \sum_d n_{dw}$.


